# How scientific literature has been evolving over the time? A novel statistical approach using tracking verbal-based methods.

Daría Micaela Hernández[a,b], Mónica Bécue-Bertaut[b], Igor Barahona[b].

[a] Centro Mexicano de Estudios Económicos y Sociales
Napoleón 54, Col. Moderna 3500, México D.F. (México)

[b] Universitat Politècnica de Catalunya
Department of Statistics and Operational Research,
Jordi Girona 1-3, 08034 Barcelona (Spain)

## Abstract

This paper provides a global vision of the scientific publications related with the Systemic Lupus Erythematosus (SLE), taking as starting point abstracts of articles. Through the time, abstracts have been evolving towards higher complexity on used terminology, which makes necessary the use of sophisticated statistical methods and answering questions including: how vocabulary is evolving through the time? Which ones are most influential articles? And which one are the articles that introduced new terms and vocabulary? To answer these, we analyze a dataset composed by 506 abstracts and downloaded from 115 different journals and cover a 18 year-period.

## 1. Introduction

New advances on science are developed on what was previously discovered. These are the foundations for introducing new progress and knowledge. At this respect, researchers have to establish differences among what is known and what is still to investigate. In this paper, a collection of abstracts is considered as what is known and thereby the starting point. This collection was created by downloading each paper (or abstract) individually from scientific databases. Later, by applying several statistical methods, it was possible to handle this large dataset and extract relevant information from it. Multidimensional methods as correspondence analysis and factorial analysis are applied for obtaining intriguing conclusions.

This paper provides a global vision of the scientific publications related with the Systemic Lupus Erythematosus (SLE), taking as starting point article abstracts. Through the time, abstracts have been evolving towards higher complexity on used terminology, which makes necessary the use of sophisticated statistical methods and answering questions including: how vocabulary is evolving through the time? Which ones are most influential articles? And which one are the articles that introduced new terms and vocabulary? To answer these, we analyze a dataset composed by 506 abstracts and downloaded from 115 different journals and cover a 18 year-period.

Moreover, it is illustrated how two types of multivariate analysis for integrating the year of publication of each abstract, and finding out changes on topics and vocabulary, are applied. The analysis presented on further pages allowed us to identify trends and changes in the vocabulary through the passing of the years, which later are presented on didactic and *easy-to-understand* visual representations. On next section materials and methods are presented. The obtained results on section 3 are discussed. Finally conclusions and further lines of research on section 4 are provided.

## 2. Materials and methods

*2.1. Data collection.*

We focus our attention on Systemic Lupus Erythematosus, known as SLE or simply "*Lupus*". Systemic Lupus Erythematosus (SLE) is a disease on which the immune system attacks cells and tissue of the body, resulting in inflammation and damage. There is not cure for SLE and its course is unpredictable. On figure 1, the most common manifestations of SLE are displayed and more information related with its symptoms on http://www.lupusuk.org.uk/ can be found.

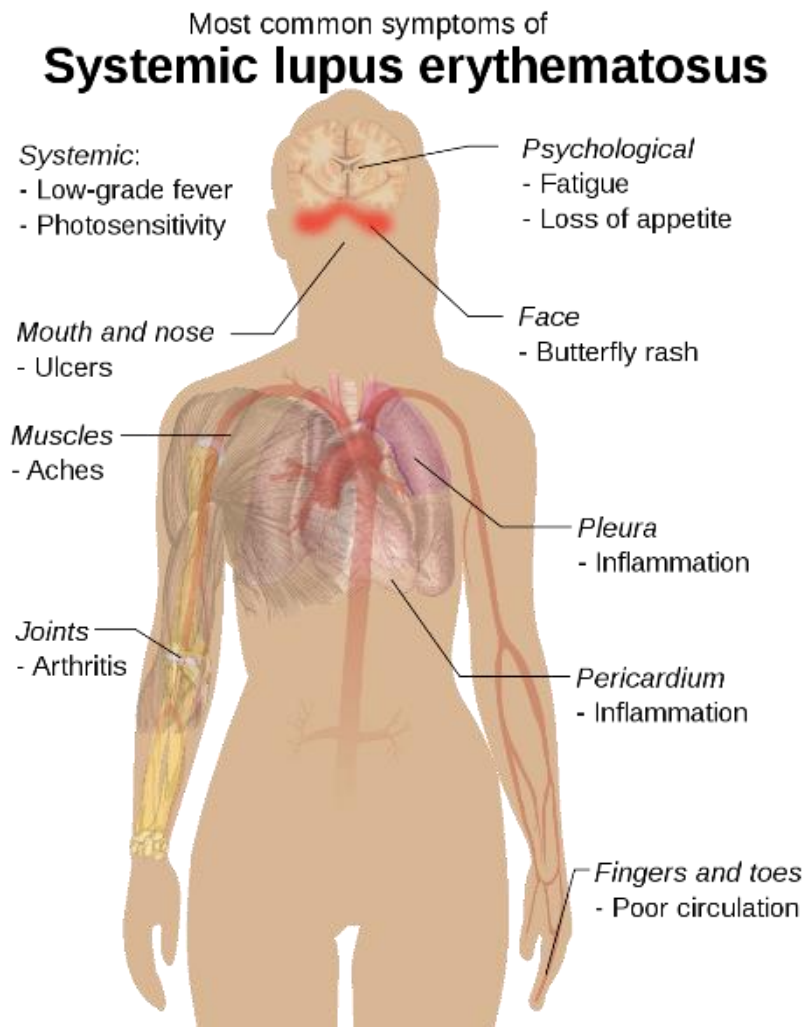

**Figure 1**. Most common manifestations of SLE

As documented below, the literature available on SLE has dramatically increased over the past 20 years. The criteria for selecting the abstracts included on the study are based on guidelines provided by the specialized SLE research group of the Hospital Clinic at Barcelona, Spain. Basically, abstracts must show two features in order to be included on this research: first, the name "SLE" should be on the title and each abstract should refer to clinical trials. Only English written documents were considered. Implementing these criteria in our search engines produced 506 abstracts which were published between January 1994 and December 2012. All documents were downloaded from MEDLINE

As additional measures of standardization, all the included abstracts should observe title, authors, journal name and year of publication. In this way, the total number of abstracts represents our object of study, and therefore we apply several statistical methods, including correspondence analysis and multiple factor analysis for contingency, on it. On Figure 2A the collected abstracts are classified by year of publication. Figure 2B shows the same abstracts classified by the journal of publication. Note that the year with the biggest number of publications is 2007, equal to 42 articles. Only eight journals have published more than 10 articles concerning this topic. Five journals, *Arthritis and Rheumatism*, *Lupus*, *The Journal of Rheumatology*, *Rheumatology Oxford, England* and *Annals of the Rheumatic Diseases*, concentrate the 60 % of the articles.

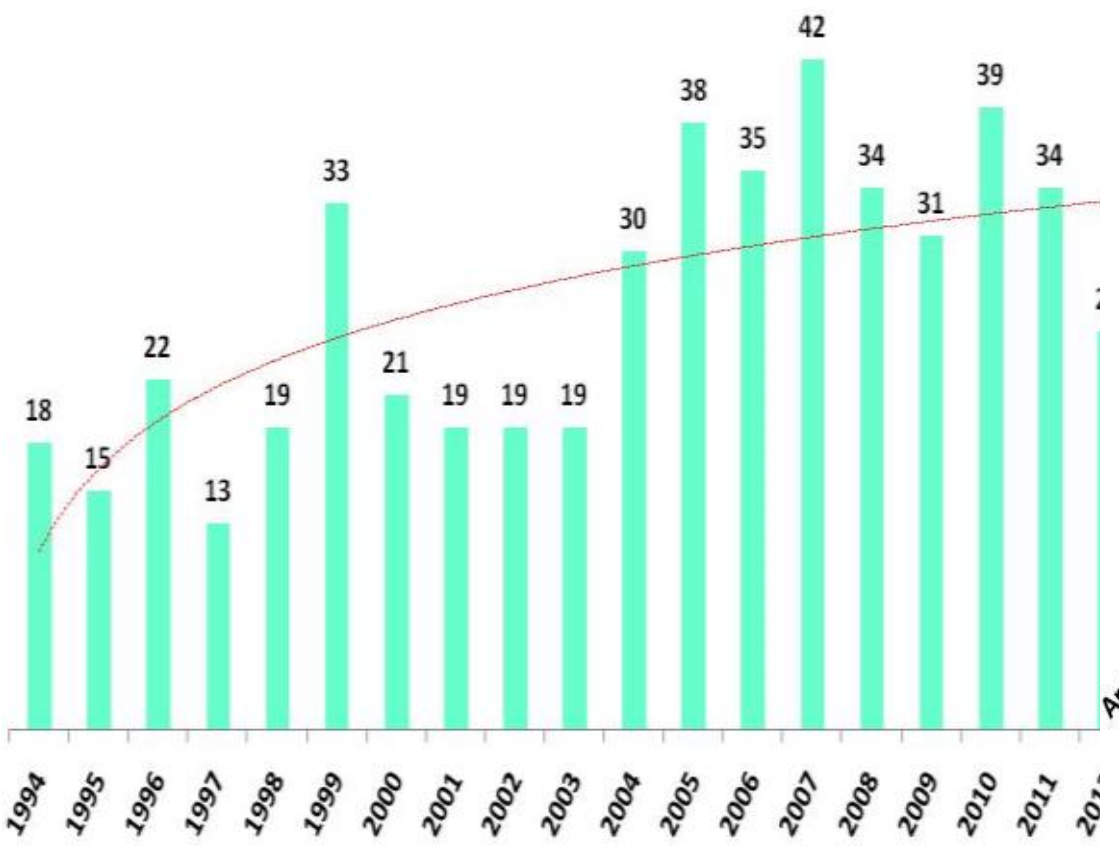

**Figure 2A.** Number of articles per year

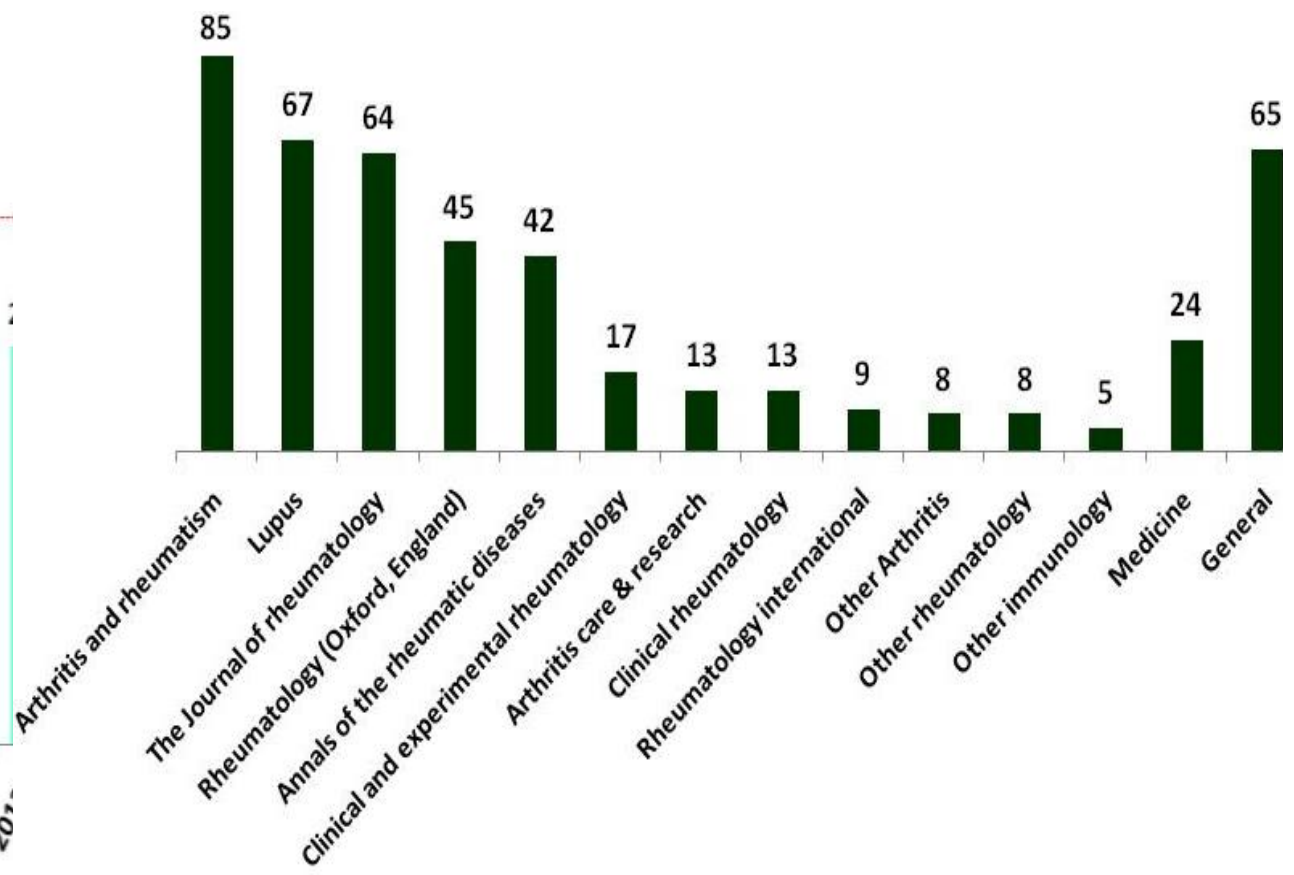

**Figure 2B**. Number of articles per journal

*2.2 Features of the dataset.*

The final corpus includes 118,094 tokens that correspond to 6,834 different words. This is an average length of 233 tokens. No lemmatization was carried out and the words were taking preserving their graphical form. Words with an occurrence of less than 5 abstracts and frequency lower than 10 times, were removed. Prepositions, conjunctions, personal pronouns and demonstratives were also removed. Figure 2 displays the structure of the corpus. It has 506 rows and 1120 columns. The rows are related to the abstracts while the columns to the words.

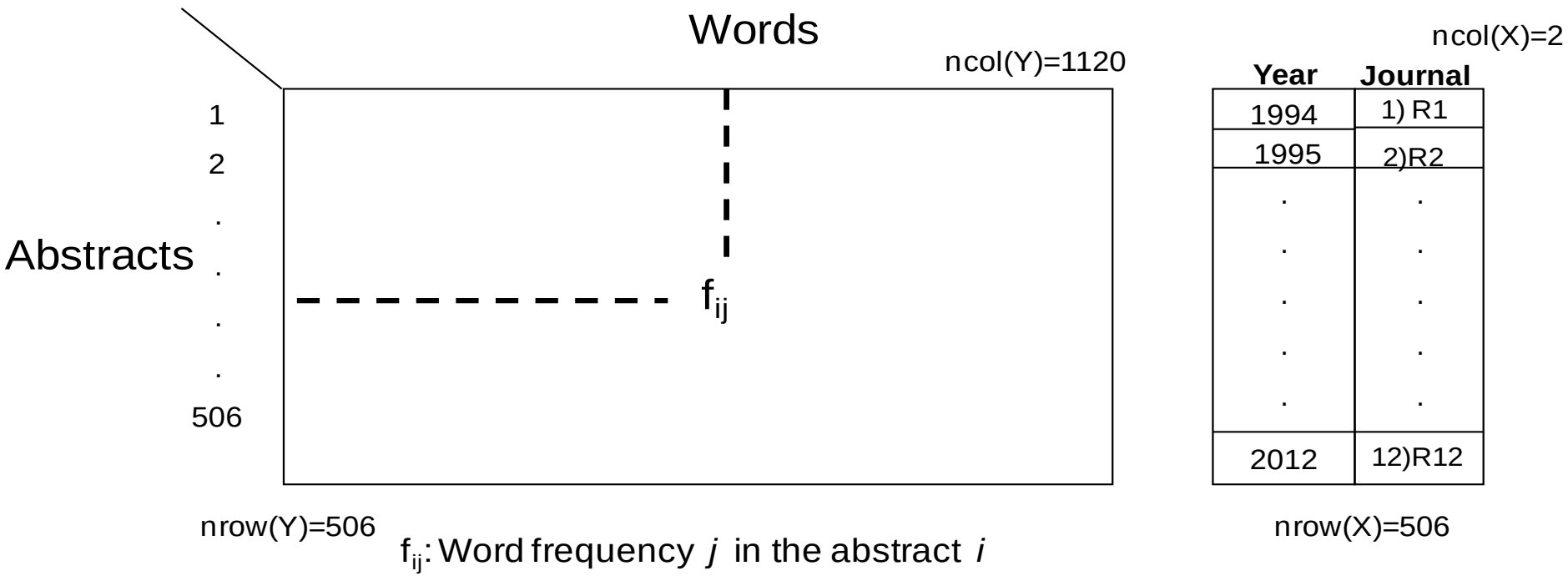

**Figure 3.** Matrix *abstracts × word.*

*2.3. Aims of the statistical analysis of the bibliography*

Content-based bibliographic studies is a discipline that attempts to detect main topics in a research domain. By making use of associations or distinctions, researchers might track changes, advances and novelties on big textual datasets. Under this perspective, counting words for topic identification makes sense because it is inferred that documents using the same words are closely related on topic and field (Benzécri, 1981). At this respect, correspondence analysis (CA) techniques (Benzécri, 1973, 1981; Lebart, L., Salem, A., & Berry, 1998; Murtagh, 2005) have proved to be a suitable tool for seeking similarities among documents by identifying words co-occurrences. The above makes the CA a useful tool to be applied on text mining (Bansard et al., 2007; Kerbaol, Bansard, & Coatrieux, 2006; Kerbaol & Bansard, 2000; Morin & Rennes, 2006; Rouillier, Bansard, & Kerbaol, 2002; Šilić, Morin, Chauchat, & DalbeloBašić, 2012).

When doing textual mining, it is identified that progress on science takes place by establishing a link among pioneer words of the first publications, which later are complemented or enriched by further authors. This progress is expressed on how the vocabulary evolves over the time, and how the introduced words are adopted on further research. To track the evolution, a time-oriented viewpoint is adopted, but the word co-occurrence approach is maintained. Multiple factor analysis for contingency tables (MFACT; (Bécue-Bertaut & Pagès, 2004, 2008) are applied on these purposes. Maps of words and abstracts, which takes into account both chronology and co-occurrences, are utilized for disclosing the evolutions on vocabulary by the passing of the years. In other words, by taking advantage of the features of MFACT for comparing chronology and vocabulary on the abstracts, novelties are spotted. MFACT is also a powerful tool for identifying abrupt changes on vocabulary or periods characterized by a flat use of words.

### *2.4. Correspondence analysis*

From a content point of view, correspondence analysis (CA) is an essential statistical method for performing text mining (Benzécri, 1981; Lebart, 1998; Murtagh, 2005). At first, the corpus has to be encoded into a lexical table, on which rows are the abstracts while words are the columns. The weights for *row-abstracts* and *column-words* are automatically handled on the CA analysis. The proportionality among the margins of the lexical table, the length of abstracts and word counts are also managed during the analysis. In similar way, rows and columns spaces, which are given by the chi-square distances, are used for computing the principal axes on each chart. As result, abstracts are mapped closer when they use a similar vocabulary, and words are mapped closer when they are more frequently present on the same abstract. This demonstrates the suitability of CA for retrieving *synonym-relationships* without introducing external information on the dataset. That is to say, abstracts whose content is closer on meaning, but expressed on different words, are plotted closer together. Additionally, CA allows us to plot elements of the corpus on a particular way, on which hidden relations among words and abstracts are uncovered. Consequently, it identifies similarities and differences between abstracts in terms of vocabulary or content (Lebart, L., Salem, A., & Berry, 1998). Basically three types of useful information is obtained from the CA.

- Similarities among abstracts based on their verbal content.
- Similarities among words based on their distribution.
- Associations among words, based on their context.
- Mutual associations among abstracts and words through a simultaneous representation of rows and columns, respectively.

Regarding with its interpretation, CA seeks, axis by axis, the "*metakeys*" and "*metadocs*" which better characterize similarities among abstracts based on their verbal content. For example, on a given axis, a "*metakey+"/"metakey" is* defined by the set of words which most contribute the inertia and simultaneously, is projected on the positive/negative part of the axis (Kerbaol et al., 2006). Therefore, two "*metakeys*" or two "*metadocs*" characterize each axis depending on the word/abstract configuration. The highest contributory word/abstracts are projected at only one axis. With this, it is evident that words belonging to the same *metakey* are frequently used on the same abstracts and, consequently, they relate to the same topic. It is to say that one particular word is used on two or more different contexts. Note that a different context is associated with different meaning. For instance, the word "*manifestations*" may be referred to the declarations that somebody makes about a particular topic. In contrast, the same word might be used as synonym of "*symptoms*" observed on the patient. Likewise, the words included on high number of *metakeys*, known as high dimensional words, provide an important weight to differentiate topics on the corpus. A "*metadoc"* and a "*metakey*" which are characterized on the same axis, are quantitatively related through CA. The abstracts that belong to the same "*metadoc*" use words associated with the same "*matekey*". In this way, abstracts related with the same topic, are rapidly clustered by identifying their associated *metakeys*.

*2.5. Multiple factor analysis for contingency tables*

According with Escofier & Pagès (2008), the Multiple Factor Analysis (MFA) offers the possibilities for introducing the chronology as active on the analysis and further mapping the abstracts based on their vocabulary. It is similar to CA, but also adding publication year.

*2.5.1. Principles*

The MFACT (Bécue-Bertaut, Álvarez-Esteban, & Pagès, 2008; Bécue-Bertaut & Pagès, 2004), is an extension of the MFA. It is able to multiple tables by juxtaposing several quantitative, categorical and frequency columns. In order to apply MFACT on this research, data is structured on the following way. The abstracts are described by both the vocabulary and the chronology. MFACT gives an active role to both vocabulary and chronology, but balances their influence in the global analysis. If a relationship does exist between chronology and vocabulary, then the given axis will be highly correlated with them. On Bécue-Bertaut et al., (2008) a detailed explanation of how MFACT are applied on text mining is provided.

*2.5.2. Types of results*

In the case of the particular structure described above, MFACT offers the following:

- Global results on the active rows and columns: eigenvalues, representations of the row-abstracts and column-words. In this global representation, the distance between two abstracts corresponds to the weighted sum of the distances induced by both the vocabulary (as in CA performed on the lexical table) and the standardized chronology (classical Euclidean distance).

- Partial results on the active rows and columns: superimposed representations of the abstracts from either the chronology or the vocabulary viewpoint, easing the discovery of those abstracts that are more advanced in the use of vocabulary (here, called pioneer works because their vocabulary is used in advanced way). Illustrative columns can be introduced and displayed on the principal graphs. Year-in-categories and journal, both considered as categorical illustrative variables, are of interest. Year-in-categories is built from the quantitative variable publication year, with the different years as the categories of this variable. Each category (year or journal) is displayed, as illustrative, at the centroid of the abstracts published during that year or journal, either on the global or partial (chronology or vocabulary) representations.

- The category-years from only the vocabulary viewpoint. The trajectory of the years is the trace of the vocabulary evolution and its rhythm, with possible steps forwards or backwards. This trajectory detects gaps in vocabulary renewal and leads to the discovery of abrupt changes and identifies lexically homogeneous periods in the corpus.

*2.5.3. Validation*

A test based on permutations assesses the significance of the first global eigenvalue that accounts for the link between chronology and vocabulary. The null hypothesis is the exchangeability of the publication year with respect to the abstracts or, equivalently, the non-existence of a chronological dimension in the vocabulary variability. Thus, the rows of the column-year are randomly permuted, for every permutation a MFACT is performed, leading to an empirical distribution of the first eigenvalue under the null hypothesis. A large, number of replications allows to obtain the p-value associated with the observed value of the first eigenvalue (one-tail test).

### *2.6. Lexical characteristics of periods*

As mentioned above, a chronological partition of the corpus is obtained by cutting the trajectory of the years from the vocabulary viewpoint (vocabulary partial representation) where large gaps between consecutive years represents substantial changes in the vocabulary. The lexical characterization of the periods facilitates the understanding of the introduced novelties. To characterize these periods, it is possible to identify the following features (Lebart, L., Salem, A., & Berry, 1998):

- The characteristic words of each period or words whose frequency in a period is significantly greater than what randomness would indicate. This observation provides evidence about the content of the abstracts of the period.
- The characteristic increments or words whose frequency significantly increases in a period compared to the previous periods. The increments denote topics that newly appear in the corpus or sharply increase their presence.
- The chronological characteristic words, which are the words assigned to the period or groups of consecutive periods that they best characterize. These words qualify the former information by showing that changes are not abrupt.

#### *2.6.1. Characteristic words*

With the aim of identify highly frequent (versus highly infrequent) words in parts of the corpus, such as those formed by grouping abstracts on a yearly basis or by journals, the test described hereafter is performed.

The following notation is used:

– *n..,* the grand total, that is, the total number of occurrences in the whole corpus;

– *n.j,* the number of occurrences in part *j*;

– *ni..,* the total count of word *i* in the whole corpus;

– *nij* the count of word *i* in part *j*.

The count *nij* of word *i* in part *j* is compared to the other counts that are obtained with all the possible samples composed of *n.j* occurrences randomly extracted from the whole corpus without replacement (which is the null hypothesis). If the word *i* is relatively more frequent in part *j* than in the whole sample, that is, if *nij/n.j>ni./n..* (less frequent in part j than in the whole sample), the p-value of the test is computed through (1) (through (2)).

1) $$p_{i,j} = \sum_{x=n_{ij}}^{n_{.j}} \frac{\binom{n_{i.}}{x}\binom{n_{..}-n_{i.}}{n_{.j}-x}}{\binom{n_{..}}{n_{.j}}}$$

2) $$p_{i,j} = \sum_{x=1}^{n_{ij}} \frac{\binom{n_{i.}}{x}\binom{n_{..}-n_{i.}}{n_{.j}-x}}{\binom{n_{..}}{n_{.j}}}$$

#### *2.6.2. Characteristic increments*

As a chronological corpus is manipulated, words whose usage significantly increases in a given period (characteristic increments) are considered as relevant information. The same test is performed for every couple (word, period), truncating the corpus at the end of the period under study.

#### *2.6.3. Chronological characteristic words*

Certain words better characterize a group of consecutive periods than a single period. Therefore, each word is successively tested as a possible characteristic word of each period (first level), of each group of two consecutive periods (second level), of each group of three consecutive periods (third level) and so on. A p-value is associated with each test. At the end of the process, each word is assigned as a chronological characteristic word to the period or group of periods that it best characterizes, i.e., to the group of periods associated with the lowest p-value (under the condition that this p-value is under 0.05). Results are detailed discussed on the following section.

## 3. Results

### *3.1. Glossary of frequent words*

Based on the glossary of the principal topics, SLE is classified in six groups: etiology, epidemiology, symptoms, prognosis, diagnostics and treatment. Note that one word might be presented in more than one topic.

| Symptoms | | Diagnostic and Prognostic | Aetiology | Drugs and/or Treatment | Epidemiology |
|---|---|---|---|---|---|
| **brain** | **renal** | biopsy | antiphospholipid | *methotrexate* | *women* |
| **pulmonary** | **nervous** | prognosis | dsdna | *cytotoxic* | *race* |
| **thrombocytopenia** | **hypertension** | prognostic | dna | *dietary* | *ethnic* |
| **chest** | **urinary** | resonance | hormonal | *cyclophosphamide* | *cohort* |
| **articular** | **cholesterol** | magnetic | atidna | *prednisone* | *african* |
| **vasculitis** | **cardiovascular** | disease | lipoprotein | *dose* | *age* |
| **blind** | **carciac** | illness | hormone | *antibodies* | *male* |
| **joint** | **glomerulonephritis** | bilag | genotype | *drug* | *hispanic* |
| **defect** | **chronic** | april | cell | *intravenous* | *hrqlo* |
| **symptoms** | **systolic** | plama | B-Cell | *hydroxychloroquine* | *ethnicity* |
| **nephropath** | **arthritis** | vascular | estrogen | *anticardiolipin* | *caucasian* |
| **abnormalities** | **depletion** | rheumatology | antibody | *therapeutic* | *gender* |
| **lung** | **inflammation** | blood | allele | *immunosuppressive* | *children* |
| **cutaneous** | **neuropsychiatric** | bmd | pathogenesis | *corticosteroid* | *multiethnic* |
| **platelet** | **bone** | calcium | snps | *chloroquine* | *ethnically* |
| **pressure** | **damage** | sledai | gene | *azathioprine* | *population* |
| **toxicity** | **rheum** | died | polymorphism | *antigens* | *person* |
| **urine** | **coronary** | evaluation | genetic | *immunoregularors* | *human* |
| **manifestations** | **thrombosis** | infection | genotiped | *rituximab* | *patient* |
| **anaemia** | **atherosclerosis** | chronic | polymerase | *belimumab* | *candidate* |
| | | mortality | haplotype | *blys* | *adult* |
| | | predictive | | *pharmacokinetics* | *case-control* |

**Figure 4.** Topics from the glossary of words

### *3.2. CA on the abstracts×words table*

The CA is performed on the table of order 506 abstracts and 1120 words. It provides the *metakeys* and *metadocs* associated with the principal axes. The first two axes, with eigenvalues equal to 0.30 and 0.27, correspond to the axes on which a high number of words and abstracts contribute. Note that they account for a minor part of the global inertia (together, 2.72%). This low percentage, typical when dealing with large spare matrices but highly significant with respect to the hypothesis of independence between words and abstracts, is frequently associated with a satisfactory structure of the data, as discussed by Lebart et al. (1998).

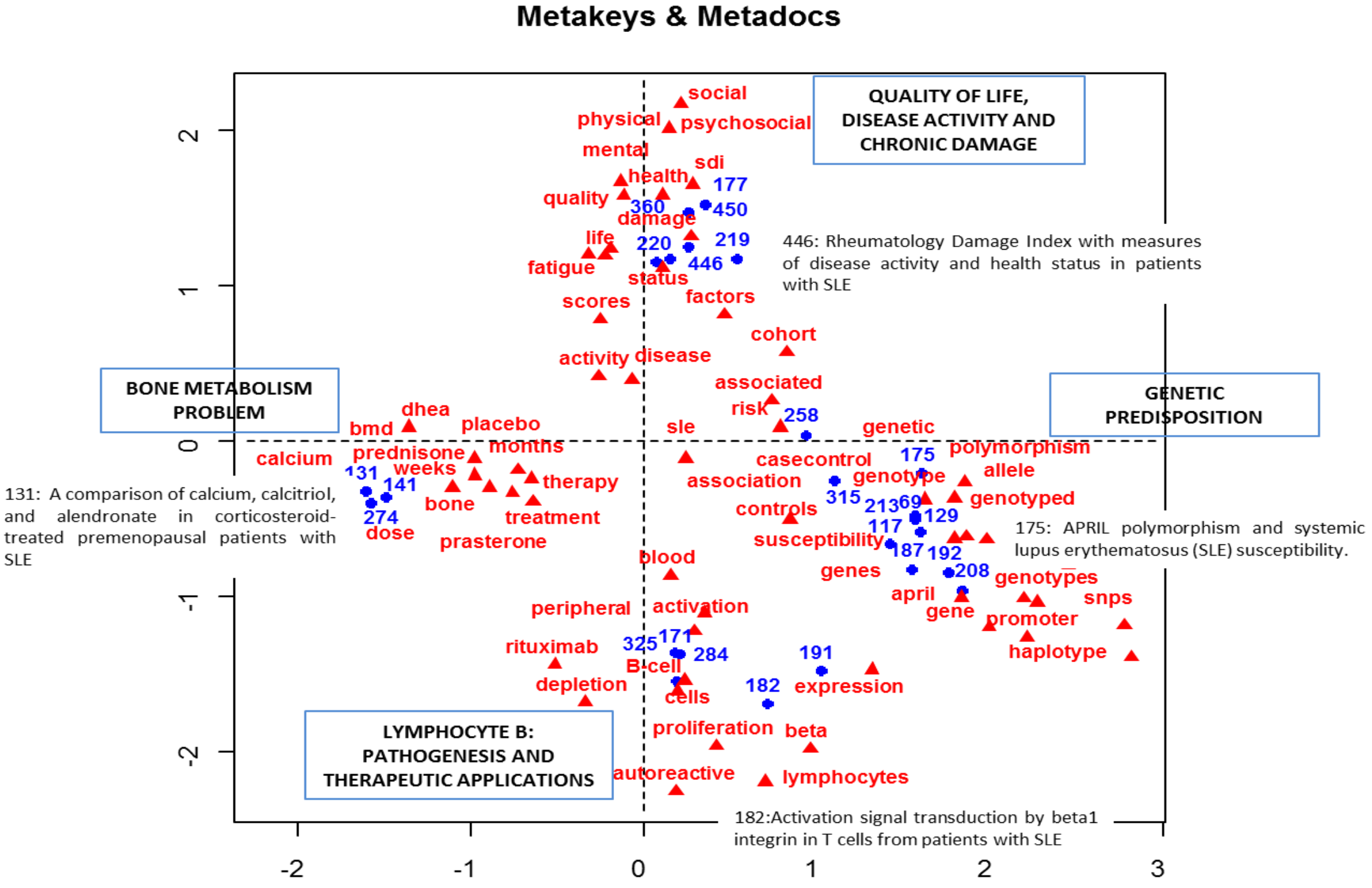

**Figure 5.** Representation of the most contributory words and abstracts on the first CA principal plane

On Figure 5 are shown the *metakeys* and *metadocs* which characterize the first principal plane. These *metakeys* and *metadocs* gather words and abstracts whose contribution is over 6 times the mean contribution of their respective set on any of the first two axes. The first axis contrasts a set of words (or *metakeys*) related to *Bone Metabolism* (BMD, praesterone, calcium, DHEA, therapy, MGDAY, prednisone, spine and lumbar) and *Genetic Tendency* (allele, gene, polymorphism, genotypes, SNPS, and haplotype). Additionally, these two *metakeys* are, in turn, opposed on the second axis (Lymphocytes B- aethiopatogeny and therapeutic applications and Life's Quality, Disease Activity and Chronic Damage). These oppositions are also found in the *metadocs*, or sets of contributory abstracts.

A summary of the information conveyed by these and the following axes (up to 50) is provided by the list of the high dimensional words, ranked from the number of dimensions on which they contribute (information not reproduced here). Although a reference is made to main first factorial plane, there are many more topics related to this issue such as treatments, quality of life, chronic damage, clinic consequences, epidemiology and new drugs. See table 1

| Topics | Metakeys | Metadocs |
|---|---|---|
| Bone Metabolism | **BMD, TREATMENT, PLACEBO ,MONTHS , PRASTERONE, CALCIUM, DHEA, THERAPY, MGDAY, DOSE, PREDNISONE, WEEKS, SPINE,** lumbar, efficacy, randomized, safety, received, month, cholesterol, mgkg, hip, reduction, cyclophosphamide, bone, response, taking, trial, density, baseline, effective, doubleblind, improvement, adverse, mineral, treated, dietary, daily, effects | **19004038, 18634158, 15801015, 15529389** |
| Genetic Tendency | **ASSOCIATION, ALLELE, GENE, POLYMORPHISM, ASSOCIATED, POLYMORPHISMS, RISK, SUSCEPTIBILITY, CONTROLS, GENOTYPES, SNPS, GENES, GENETIC, SLE, HAPLOTYPE, EXPRESSION, COHORT, GENOTYPED, CASECONTROL**, april, genotype, factors ,variants, independent, odds, found, functional, promoter, alleles, fcgammariia, japanese, diagnosis, healthy, population, associations, linkage, presence, damage, ethnicity, confidence, vascular, seizures, hla, cerebrovascular. | **20498198, 19019891, 17569747, 17307753, 17393452, 17028114, 17092253** |
| Lymphocytes B-aethiopatogeny and therapeutic applications | **CELLS, CELL, EXPRESSION, LYMPHOCYTES, DEPLETION, GENE, RITUXIMAB, BLOOD, BETA, ACTIVATION,** polymorphisms, susceptibility, peripheral, normal, fcgammariia, dna, autoreactive, controls, polymorphism, proliferation, genotypes, haplotype, healthy, naive, serum, allele, autoimmune, immune, antibodies, genes, lymphocyte, snps, association, anticd, antibody, april, mab, promoter, vitro, flow. | **17469136, 17312177, 15529346, 12571855** |
| Quality of life, chronic damage | **DAMAGE, HEALTH, DISEASE, PHYSICAL, SDI, ACTIVITY, SOCIAL, MENTAL, FATIGUE, SCORES , QUALITY, LIFE, FACTORS, PSYCHOSOCIAL, STATUS, PSYCHOLOGICAL,** accrual, duration, costs, outcomes, variables, hrqol, mcs, index, insurance , score, canada, international, college, outcome , sliccacr, visits, overall, collaborating, age, diagnosis, selfreported, socioeconomic, race, american, impact, pcs, ethnicity, clinicsamerican, texas, helplessness, slamr, bilag, domains, multivariable, care, african, support, hispanics, analyses. | **19327231, 17551377, 16880189, 16905579, 15757968, 11414264, 9034988, 9008599** |
| clinic consequences | **SPECT, PERFUSION, DEFECT, BRAIN, MRI, NPSLE, ABNORMALITIES, DIFFUSE, IMAGING, CEREBRAL, PATIENTS, CEREBROVASCULAR, MORTALITY, SYNDROME, CNS, MAGNETIC , RESONANCE, CARDIAC, PULMONARY, INVOLVEMENT,** manifestations, diagnostic, positive, psychosis, myocardial, neuropsychiatric, survival, persistent, abnormal, causes, detected, nervous, death, cyclophosphamide, underwent, symptomatic, deaths, useful, central, sensitive, lung, anticoagulant. | **16342090, 15950567, 14659847, 1160073, 10823333, 9049450, 7980671, 7562753** |
| Epidemiology and genetic | **CALCIUM, BMD, SPINE, GENE, POLYMORPHISM, ALLELE, POLYMORPHISMS, PRASTERONE, SNPS, HAPLOTYPE, APRIL,** genotypes, dhea, hip, susceptibility, lumbar, association ,expression, density, mineral, genes, premenopausal, placebo, fcgammariia, casecontrol, functional, bone, promoter, cholesterol. | **19004038, 17307753, 19004038, 20964867, 1915112, 18634158 16342090, 16320346, 15801015, 15529389, 7562753** |
| Cardiovascular Risk | **CALCIUM, BMD, SPECT, SPINE, LUMBAR, PERFUSION, HIP, BONE, PRASTERONE, PREMENOPAUSAL, DEFECT , MINERAL, TAKING, DENSITY, RISK,** brain, mri, loss, cardiac, imaging, cholesterol, glucocorticoids, women, pulmonary, abnormalities, hrt, postmenopausal, sle, dhea, diffuse, myocardial, increases, receiving, controls, coronary, significant, artery, dietary.. | **18372357,15797975, 15593203** |

**Table 1.** Representation of the *metakeys* (the most contributory words) and *metadocs* (the most contributory abstracts) on the 10 firsts CA axes.

## 3.3. Chronological evolution

### 3.3.1. MFACT: global analysis on the first plane

MFACT is performed on the multiple table by juxtaposing the *abstracts*×*words* table (with dimensions 506 ×1120) and the *abstracts*×*year* column (with dimensions 506×1), leading to map the abstracts according to both their vocabulary and chronology. The first eigenvalue of MFACT is equal to 1.34. This value is far from randomness, as evidenced by the permutation test (p-value = 0.000 computed from 1000 replications) and is a medium value; in this case, the maximum is equal to 2 (Escofier, B., & Pagès, 1992). The first axis on which the words contribute 33.4% and the year contributes 66.6% of the inertia is a dispersion direction, present in both sets of columns. Words and chronology, that differs from the first principal axis obtained in the separate analysis of the words (that is, in CA). With regard to the words, MFACT conserves a small proportion of the total inertia on the first axes as CA applied to short texts. However, because the first MFACT axis retains a proportion of inertia of the word cloud relatively close to the proportion of inertia explained by the first CA axis. This can be considered as a relevant dispersion direction for the cloud. Regarding to chronology, the high correlation of the MFACT first axis with publication year (corr = 0.94) suggests that this axis accounts for the variability of the vocabulary related to time.

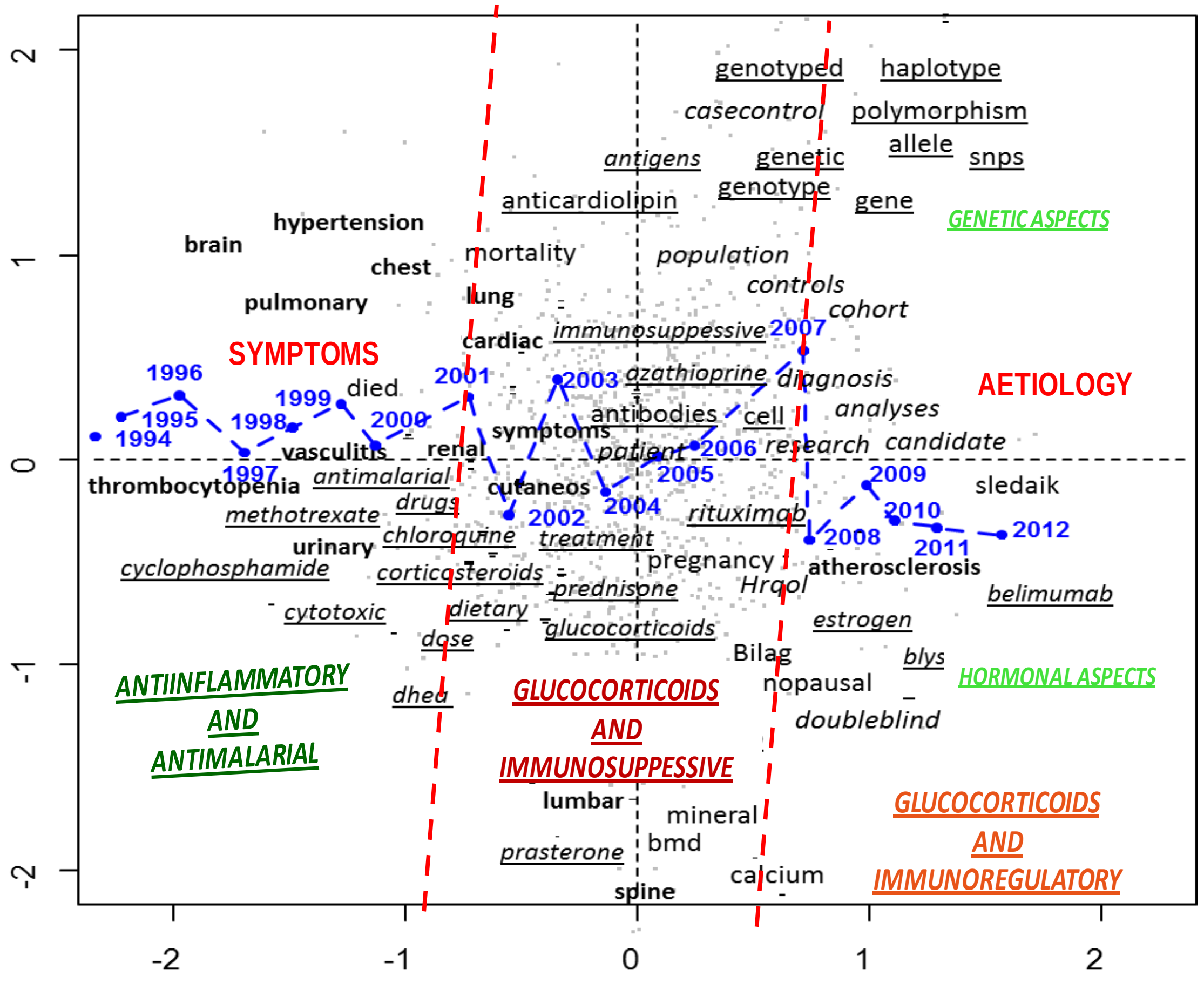

**Figure 6.** Global representation of an excerpt of the words with representation of years_in _categories (illustrative) on the first MFACT principal plane.

Figure 6 shows the global representation of the words and abstracts on the first principal plane. In this figure, the abstracts are differentiated by three time periods (1994–2000; 2001–2006; 2007–2012). These time divisions are justified in Section 2.5.2. This first axis opposes words related to etiology from two different points of view: genetic aspects and hormonal aspects (on the positive side) and symptoms and drugs (on the negative side). From the quantitative variable year, the categorical variable ‘‘*year_in_categories*’’ is created. The categories-years (or group of years) are projected as illustrative on the first axis at the centroid of the abstracts published during this period, from the global point of view. The most relevant words are also represented on this axis from their coordinates. This representation allows for the replacement of vocabulary renewal into time flow. As expected, the categories-years lie on the first axis in their natural order, although they are separated by different length intervals.

*3.4. Segmentation homogeneous periods*

The segmentation suggested on figure 7 and commented on in the former section is retained. The periods 1994–2000, 2001–2006 and 2007–2012 are described by their lexical features, enabling the characterization of the evolution of the vocabulary. The results are consistent with those illustrated on figure 6. Changes in lexicon are due mainly to clinic symptoms, their etiology and the presence of new developed medications.

**1994-2000**

**Chronological characteristic words**

manifestations, pulmonary, mgd, prednisone, cytotoxic, patients, ana, brain, dhea, pressure, hypertension, excretion, mortality, antiphospholipid, psychosocial, cutaneous, slam, discontinued, psychosocial, lung, antiphospholipid, thrombocytopenia, vasculitis, urinary, articular, nervous, treatment, drug, urine, mortality, cyclophosphamide, nephropathy, anemia, antimalarials.

**2001-2006**

**Chronological characteristic words**

depletion, women, perfusion, antibody, social, fever, children, apoptosis, chloroquine, prasterone, genotypes, defect, symptomatic, rituximab, double, cause, exercise, hispanics, immunological, intervention, adult, aged, cell, cholesterol, africanamericans, rheumatology, control .

**Characteristic increments**

allele, antibody, costs, depletion, events, flare, genes, growth, hispanics, journal, mmf, placebo, pmi, polymorphisms, prasterone, rituximab, support, titres visits , weeks, women

**2007-2012**

**Chronological characteristic words**

belimumab, risk, csle, nih, snps, blys, bilag, sledaik atherosclerosis, calcium, hrqol ,aza, juvenile ,selenasledai, signals, ratio, events, signaling, cohort, analysis, registration, neuropsychological, recombinant, haplotype, pharmacokinetics, gender, worsening, estrogen, cardiovascular, vitamin, gene

**Characteristic increments**

Accrual, allele, anaemia, analysis ,april, ,associated, association, autoimmunity, beta, bilag ,cancer, casecontrol, cells, cohort, genotyped, haplotype, ,increased, japanese, lymphocytes, mrna, polymorphism, ,risk, belimumab, bilag, biological, blys, calcium, candidate, cells, cognitive, diagnosed, differential, expression, flare, gday, healthrelated, hhsunited, hrqol, igg, iuml, ,juvenile, pharmacokinetics, plasma.

**1994-2006**

**Chronological characteristic words**

Manifestations, pulmonary, dhea, months, patients, brain, antibodies, cytotoxic, psychosocial, cyclophosphamide, thrombocytopenia, antiphospholipid, prednisone, died, treatment, cutaneous, vasculitis, dietary, abnormalities, corticosteroids, nervous, clinical, urine, mortality, monthly, antimalarials, anticoagulant, lung, European, drug, pregnancy, intervention, blind, renal, causes, infections

**2001-2013**

**Chronological characteristic words**

Belimumab, csle, risk, blys, sledaik, hrqol, bilag, cells, juvenile, mgkg, atherosclerosis, calcium, Japanese, polymorphism, genotyped, analysis, gene, neuropsychological, allele, cohorts, vitamin, estrogen, rheumatologists, casecontrol, haplotype, gday, nucleotide, multiethnic, cardiovascular, candidate, genetic, ethnicity, biological, genes, rituximab, double-blind

**Figure 7.** Lexical chronological characteristics and significant lexical increments.

Figure 6 shows how the studies in medical sciences related to SLE have diversified from 1994 to 2012. In the 1994–2000 period, the main issue was "*symptomatoly*" and the treatments were almost exclusively based on "*anti-malaric*" (chloroquine and hydrochloroquine) and "*antiinflammatory*" (methotrexate and cyclophosphamide) drugs. Both 2001 and 2002 introduced many new words without remarkable changes until 2006, when novelties in genetic and etiology were introduced, as well as the use of glucocorticoids (prednisone) and immune-supressive (azathropine and prasterone) as partial treatment. In the 2007–2012 period, research focused on genetic and hormonal aspects. Some biological therapies have been used (*rituximab* and *belimumab*) in this period of time.

The words introduced in 2007 showed an increasing interest for the researchers. This year some cohort and case-control were performed and new issues about etiology were also studied. Although topics of interest prior to 2001 remain in the following years, there is a more noticeable homogeneity in the vocabulary observed from 2001–2012 than from 1994–2006, as it is shown by the greater number of chronologically characteristic words relative to 2001–2012 in comparison to 1994–2006. This observation is reinforced by the strong similarity between the lexical increments observed in 2001–2006 and 2007–2012. For instance, Belimumab appears from 2001–2006 (2 citations) and its usage increases from 2007–2012 (55). The words gene(s) are used 4 times from 1994 to 2000, 42 times from 2001 to 2006 and 70 times from 2007 to 2012; The words women, in some cases named female(s), are used 55 times from 1994 to 2000, 93 times from 2001 to 2006 and 127 times from 2007 to 2012, Nowadays, most research has been focused on hormones and their action on bone metabolism, especially on post-menopausal women. Last advance has been related to a new medication (Belimumab) who has an immune-suppressor and immune-regulator activity on *B-lymphocytes*. It seems that future research will be focused on drugs with similar activity. The methodology becomes more complex and requires more sophisticated statistical methods, frequently imported from other fields

### *3.4. Pioneer abstracts.*

The superimposed representation of the partial *points-abstracts* from either the chronology or vocabulary viewpoint allows for spotting the abstracts with a vocabulary ahead of their date. However, the set of words contributes to only 33.4% of the inertia of the first axis, whereas the column-year contributes 66.6%. This unequal contribution of both sets of columns to the first axis inertia makes the partial points ‘‘*chronology*’’ placed on average farther from the centroid on the first axis than the partial points ‘‘*vocabulary*’’. This is particularly clears the oldest dated abstracts due to their low numbers in the same year. These abstracts tend to mechanically present a partial point ‘‘*vocabulary*’’ ahead of the partial point ‘‘*chronology*’’. On the contrary, this effect, notably less marked, works on the opposite direction in the case of the recent abstracts, which allows for interpreting partial points ‘‘*vocabulary*’’ ahead of partial points ‘‘*chronology*’’ as corresponding to abstracts using an innovative vocabulary. These abstracts contain words with a high coordinate on the first axis of MFACT and use these words at an early stage (See figure 8).

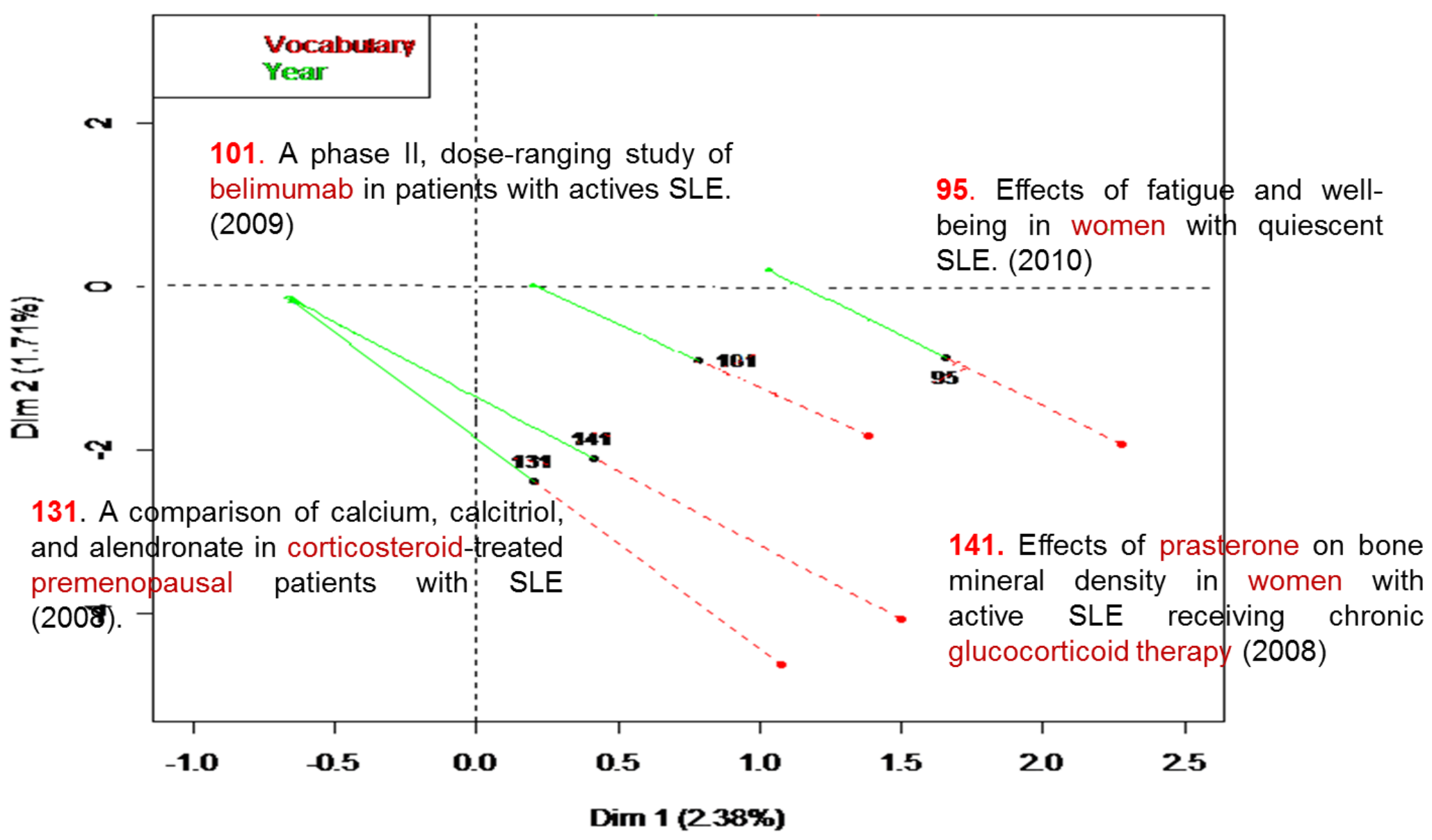

**Figure 8**. Representation pioneers abstracts on the first MFACT principal plane from 2006 to 2010.

Table 2 gives the title, first author, journal and year of the most ahead-of-its-date abstracts in 2008, 2009 and 2010, although all should be consulted. Examining their content facilitates the understanding of the research advances. The ahead-of-date abstracts often correspond to innovations. A gap in the number of ahead-of-date abstracts (1 in 2010, 2 in 2009, 8 in 2008,) corresponds to a gap in the trajectory of the years from the viewpoint of vocabulary (See figure 8). The ahead-of-date abstracts often correspond to innovations. The most recent abstracts cannot be selected from this criterion (no pioneer work has been identified on 2011 and 2012), and recent innovative articles may have been missed. Nevertheless, it is possible to look for recent abstracts (published either in 2011 or 2012) that favored advanced vocabulary. They are detectable because they present a high partial coordinate on the first axis from the vocabulary viewpoint.

| Nº | Title | First author | Journal | Year |
|---|---|---|---|---|
| 95 | Effects of dehydroepiandrosterone on fatigue and well-being in women with quiescent systemic lupus erythematosus: a randomised controlled trial. | Hartkamp A | Annals of the rheumatic diseases | 2010 |
| 101 | A phase II, randomized, double-blind, placebo-controlled, dose-ranging study of belimumab in patients with active systemic lupus erythematosus. | Wallace DJ | Arthritis and rheumatism | 2009 |
| 131 | A comparison of calcium, calcitriol, and alendronate in corticosteroid-treated premenopausal patients with systemic lupus erythematosus. | Yeap SS | The Journal of rheumatology | 2008 |
| 141 | Effects of prasterone on bone mineral density in women with active systemic lupus erythematosus receiving chronic glucocorticoid therapy. | Sanchez-Guerrero J | The Journal of rheumatology | 2008 |

**Table 2:** Excerpt of the ahead-of –date abstracts from 2008, 2009 and 2010 (Hartkamp et al., 2010; Sánchez-Guerrero J et al., 2008; Yeap SS et al., 2008)

## 4. Discussion

In this study were investigated the approaches regarding with the study of a *words-table*, which is a special 2-way contingency table. Specifically, in this *words-table* each cell is found our unit of analysis: word, keyword or lemma. The main purpose in this analysis is to obtain pertinent information from the bibliographic databases by doing text mining.

The correspondence analysis demonstrated to be a suitable statistical technique to process this sort of data for discovering significant groups of words. The multi factor analysis with contingency tables (MFACT) was also suitable for discovering trends through the passing of the years and documents pioneers.

At the end this research provided evidence of the flexibility of these methods, which can be used as strategic tools for following impact and trends in practically any scientific bibliography, regardless of its topic.